\documentclass[10pt,twocolumn,letterpaper]{article}

\usepackage{iccv}              
\usepackage{multirow}
\usepackage{float}
\usepackage{stfloats}

\definecolor{iccvblue}{rgb}{0.21,0.49,0.74}
\usepackage[pagebackref,breaklinks,colorlinks,allcolors=iccvblue]{hyperref}

\def\paperID{2} 
\def\confName{ICCV}
\def\confYear{2025}

\newcommand{\mcircle}[1]{\raisebox{1pt}{\textcircled{\raisebox{-.9pt} {#1}}}}
\newlength\savewidth

\title{\textsc{VidMP3}: Video Editing by Representing Motion with Pose and Position Priors\vspace{-2mm}}

\author{Sandeep Mishra\\
University of Texas at Austin\\
{\tt\small sandy.mishra@utexas.edu}
\and
Oindrila Saha\\
University of Massachusetts Amherst\\
{\tt\small osaha@umass.edu}
\and
Alan C. Bovik\\
University of Texas at Austin\\
{\tt\small bovik@ece.utexas.edu}
}

\begin{document}

\twocolumn[{%
\renewcommand\twocolumn[1][]{#1}%
\vspace{-3mm}
\maketitle
\begin{center}
    \centering
    \captionsetup{type=figure}
    \vspace{-8mm}\includegraphics[width=0.95\linewidth]{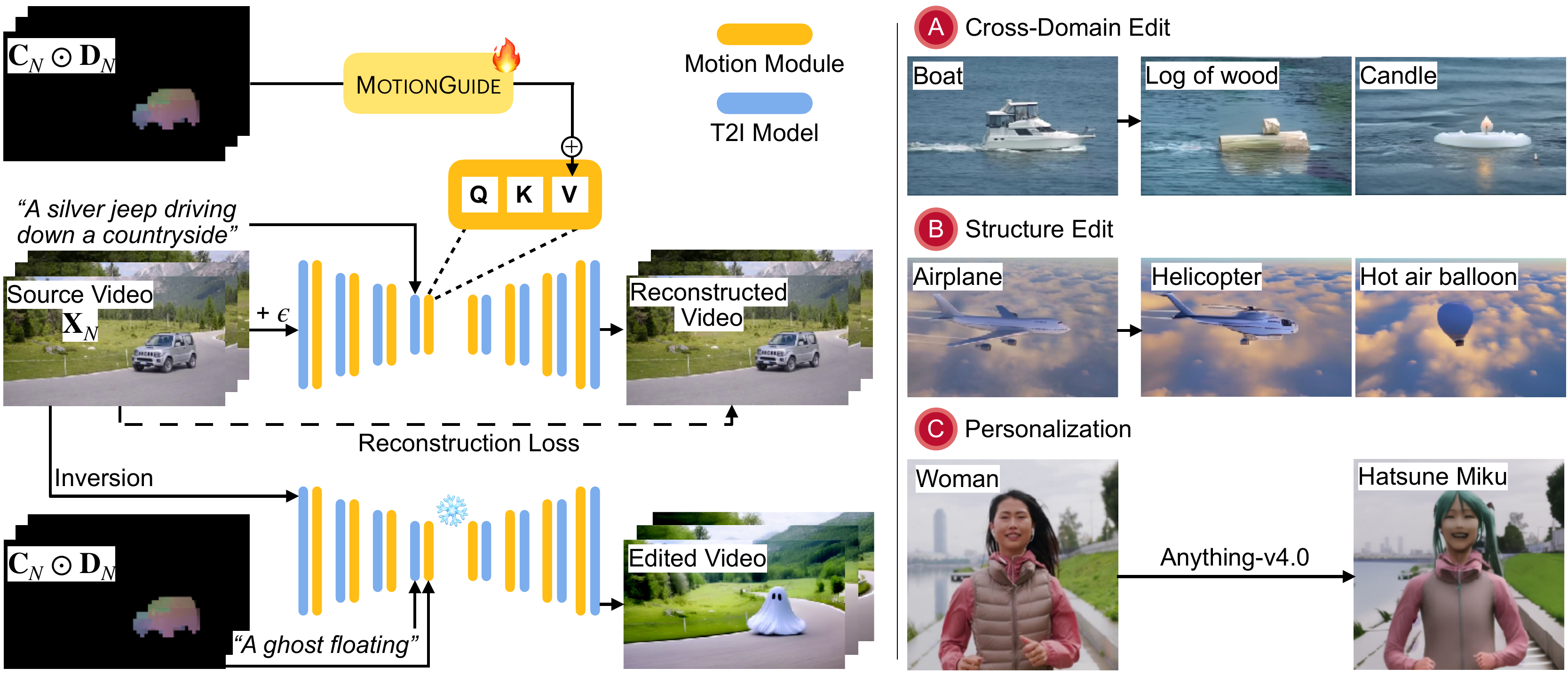}
    \captionof{figure}{\textbf{\textsc{VidMP3.}} We present a novel video editing technique that can perform challenging video editing tasks guided by pose and position priors. We introduce a \textsc{MotionGuide} module that learns a generalized motion representation from correspondence and depth maps. We inject the features of this module to the ``Value"s of the temporal self-attention layer of a T2V initialized with a T2I model. During inference, we use the correspondence and depth maps of the source video to generate a novel motion-preserved video. \textsc{VidMP3} enables the generation of challenging edits, including \mcircle{A} Cross-Domain editing, where objects with vastly different semantic meanings can be generated, \mcircle{B} Structure editing, where structure of the object can be changed significantly, and \mcircle{C} adaptation to various T2V editing tasks such as personalized editing.
    }
    \label{fig:vidmp3}
\end{center}%
}]
\vspace{-5mm}

\begin{abstract}
\vspace{-9mm}

Motion-preserved video editing is crucial for creators, particularly in scenarios that demand flexibility in both the structure and semantics of swapped objects. Despite its potential, this area remains underexplored. Existing diffusion-based editing methods excel in structure-preserving tasks, using dense guidance signals to ensure content integrity. While some recent methods attempt to address structure-variable editing, they often suffer from issues such as temporal inconsistency, subject identity drift, and the need for human intervention. To address these challenges, we introduce \textsc{\textbf{VidMP3}}, a novel approach that leverages pose and position priors to learn a generalized motion representation from source videos. Our method enables the generation of new videos that maintain the original motion while allowing for structural and semantic flexibility. Both qualitative and quantitative evaluations demonstrate the superiority of our approach over existing methods. The code will be made publicly available at \href{https://github.com/sandeep-sm/VidMP3}{this https URL}.


\end{abstract}




\section{Introduction}
\label{sec:intro}
The strong generation capabilities of text-to-image (T2I) diffusion models have encouraged the adoption of these models for video generation and editing tasks, owing to the simple architectural changes required over T2I models to enable them to generate videos. Inclusion of temporal self-attention layers and inflating 2D convolutions to pseudo 3D convolutions facilitates the generation of videos conditioned on text. While some approaches train text-to-video (T2V) models on large-scale text-video paired datasets~\cite{ho2022video, ho2022imagen, singer2022make, zhou2022magicvideo, blattmann2023align}, others explore a more data-efficient technique. These methods~\cite{wu2023tune, gu2024videoswap, zhang2024camel, song2025save} train a T2V model on a single video and use the learned priors to generate novel videos using edited text prompts. T2I models have also been used for zero-shot video editing~\cite{liu2024video, geyer2023tokenflow, qi2023fatezero, couairon2023videdit, chai2023stablevideo} by utilizing structure from a specific source video. 

Generative video editing is a task of remarkable interest to creators which enables them to create novel videos which can borrow information from a captured real video. One of the most important and under-explored sub-areas is where only motion is preserved from a source video and mimicked to generated a new video. This is the most general use-case of generative video editing, whereby the motion of the subject in the source video is preserved but structure, appearance, and semantics remain modifiable. Apart from the clear benefits of reducing costs and time for video creators, this serves an important case where a creator would want to imitate the motion of a real subject and transfer it to subjects that might be hard to capture following that specific motion e.g., imaginary concepts following the motion in a real video. 

In a data efficient setting where we want to use only a single source video to generate an edited novel video, changing the structure and domain of the subject has been a challenging task. Zero-shot video editing techniques heavily rely on the structure of the source video, and are thus unable to deviate much from the source concept. One-shot tuning techniques have shown sufficient promise, but struggle with either shape leakage, quality issues, or fail in cases of cross-domain editing. This can be attributed to unconstrained optimization over the source video~\cite{wu2023tune} or too sparse external control~\cite{gu2024videoswap}. 

We embark on learning a generalized motion representation that distentangles spatial properties of subjects from their motion. Motion of subjects is perceived by humans as the combination of their position in a 3D space and their internal pose. Thus, we choose to inject an external representation learned from pose and position priors to guide the T2I diffusion model. We hypothesize that motion can be represented as a combination of spatial correspondence maps, depth maps and 2D positional encodings. The correspondence maps provide signals for the internal pose variation of a subject over video frames, while the depth maps and positional encoding signify the 3D positions of the subject in each frame. We introduce a novel \textsc{MotionGuide} module which utilizes these maps to learn a generalized representation of motion. First, we show a proof of concept where \textsc{MotionGuide} can be used to learn the 3D trajectory and rotations of a simple moving cube. We show that the learned module is invariant to shape changes of the object but sensitive to motion changes. This shows that this module can be effectively used to induce motion-preservation with variations in shape when appropriately injected into a T2V diffusion model initialized with a T2I model. We present \textsc{VidMP3} where we inject the spatially pooled features of \textsc{MotionGuide} into the ``Value"s of the temporal self-attention layers of the T2V model. Essentially, this allows the model to understand added context in frame-to-frame correspondence, thus boosting temporal consistency. We show that \textsc{VidMP3} robustly edits subjects with significant structure and semantic shift from the subject in the source video. We also scale our method to Stable-Diffusion-XL~\cite{podell2023sdxl}, which has not been explored previously for video editing. We show that we are able to generate more diverse concepts with \textsc{VidMP3} SD-XL. In summary, our contributions are as follows:
\begin{itemize}
\item{A \textsc{MotionGuide} module that learns generalized motion representations from pose and position priors}
\item{\textsc{VidMP3}, which utilizes the \textsc{MotionGuide} module to inject external guidance to the ``Value"s of the temporal self-attention module}
\item{Adaptation to various T2I diffusion models including scaling to SD-XL.}
\end{itemize}

\section{Related Work}
\label{sec:related work}
Diffusion models have been extensively explored for video editing due to their strong generation capability and ability to conform to various kinds of conditions. Previous video editing techniques can be classified into two general categories: 1) Structure-preserved Video Editing, and 2) Motion-preserved Video Editing. We discuss prior work in these two domains in detail below.

\subsection{Structure-preserved Video Editing}
These techniques aim to edit the video while preserving structural information from the original video by relying on various cues such as depth, edge, optical flow, or attention map information.
Gen-1~\cite{esser2023structure}, Ground-a-video~\cite{jeong2023ground}, and RAVE~\cite{kara2024rave} utilize depth maps for guidance, while CCEdit~\cite{feng2024ccedit}, ControlVideo~\cite{zhang2023controlvideo}, and MAskINT~\cite{ma2024maskint} extend to the use of various controls including depth, boundary, and line drawing. MoCa~\cite{yan2023motion}, Rerender A Video~\cite{yang2023rerender}, and FlowVid~\cite{liang2024flowvid} use optical flow as guidance. VideoP2P~\cite{liu2024video}, FateZero~\cite{qi2023fatezero}, Vid2Vid-Zero~\cite{wang2023zero}, and Edit-A-Video~\cite{shin2024edit} inject attention map information from the original video while denoising the edited video. TokenFlow~\cite{geyer2023tokenflow}, COVE~\cite{wang2024cove} and DreamMotion~\cite{jeong2024dreammotion} use dense spatial correspondences among frames to ensure consistency. VidTome~\cite{li2024vidtome} develops a method that uses any of the above discussed types of guidance techniques. Codef~\cite{ouyang2024codef}, VidEdit~\cite{couairon2023videdit}, and StableVideo~\cite{chai2023stablevideo} learn a canonical representation of the video. Editing this representation allows high temporal consistency, but restricts changes in low-level features. In contrast to these methods, \textsc{VidMP3} allows significant structural and semantic changes in the subject of the given source video.

\subsection{Motion-preserved Video Editing} 
These methods aim to extract the motion from the source video while allowing significant structural changes in the edited video generated with the same motion. 

\textbf{One-shot tuning.}
Tune-a-video~\cite{wu2023tune} attaches a motion module to a pre-trained T2I model, and introduces sparse causal self-attention which uses features from other frames to compute self-attention on each frame. Tune-A-Video overfits the motion module to a single video, which is then used to generate novel videos at test-time. We find that Tune-a-Video suffers from severe structure leakage and temporal inconsistency, due to unconstrained training of the motion module on the input video. 

VideoSwap~\cite{gu2024videoswap} alleviates structure leakage by injecting keypoint correspondence information and keeping the motion module frozen. However, VideoSwap requires human effort in selecting or editing the keypoint positions. For cases which require significant size changes, VideoSwap creates a Layered Neural Atlas~\cite{kasten2021layered} of the video, in which the user is required to make desired edits. Training this LNA is significantly time consuming. Additionally, as a result of using keypoint correspondence, VideoSwap is ineffective at swapping semantically different objects. By contrast, \textsc{VidMP3} is able to swap objects with considerable structure and semantic variation, due to injecting a generalized representation of external pose and position guidance. Most importantly, \textsc{VidMP3} relies neither on human effort nor the time-intensive LNA creation process.

SAVE~\cite{song2025save} aims to disentangle the structure and motion of a subject by using a motion prompt that focuses on moving areas, but suffers from temporal inconsistencies due to leakage in areas surrounding the moving object, as evidenced in their results. CAMEL~\cite{zhang2024camel} injects motion prompts into the temporal attention module, which is then learned from the video. By contrast, our method uses external pose and position guidance to learn a more consistent representation of motion.

Emu-Video~\cite{girdhar2023emu} attaches an image editing and video generation adapter over a pre-trained T2I model, which is then tuned on a dataset of several videos. \textsc{VidMP3} instead extracts various kinds of information from a single video to generate a novel edited video.

\textbf{Pose-guided video editing.} 2D/3D pose-guided video editing has been explored specifically for humans and human-like entities in Follow-Your-Pose~\cite{ma2024follow}, DreamPose~\cite{karras2023dreampose}, DeCo~\cite{zhong2025deco}, MagicPose~\cite{chang2023magicpose}, MagicAnimate~\cite{xu2024magicanimate}, AnimateAnyone~\cite{hu2024animate}, EVA~\cite{yang2024eva}, and DynVideo-E~\cite{liu2024dynvideo}. \textsc{VidMP3} instead explores pose-guided editing in a more general context with pose being represented using correspondence maps. This representation allows us to generate subjects which are highly semantically and structurally different from the subject in the source video, while accurately following the motion of the source video.

\textbf{Propagation from first frame editing.}
AnyV2V~\cite{ku2024anyv2v} and I2VEdit~\cite{ouyang2024i2vedit} use a separate model for editing the first frame of the video and then propagate the edit to the other frames. While these methods can significantly change the structure of the subject, they are limited by the image-editing technique they utilize. AnyV2V suffers from severe temporal inconsistencies when modeling videos with significant motion (see Appendix). \textsc{VidMP3} instead learns the motion representation from the source video and jointly models it across frames.

\begin{figure}
    \centering
    \includegraphics[width=0.93\linewidth]{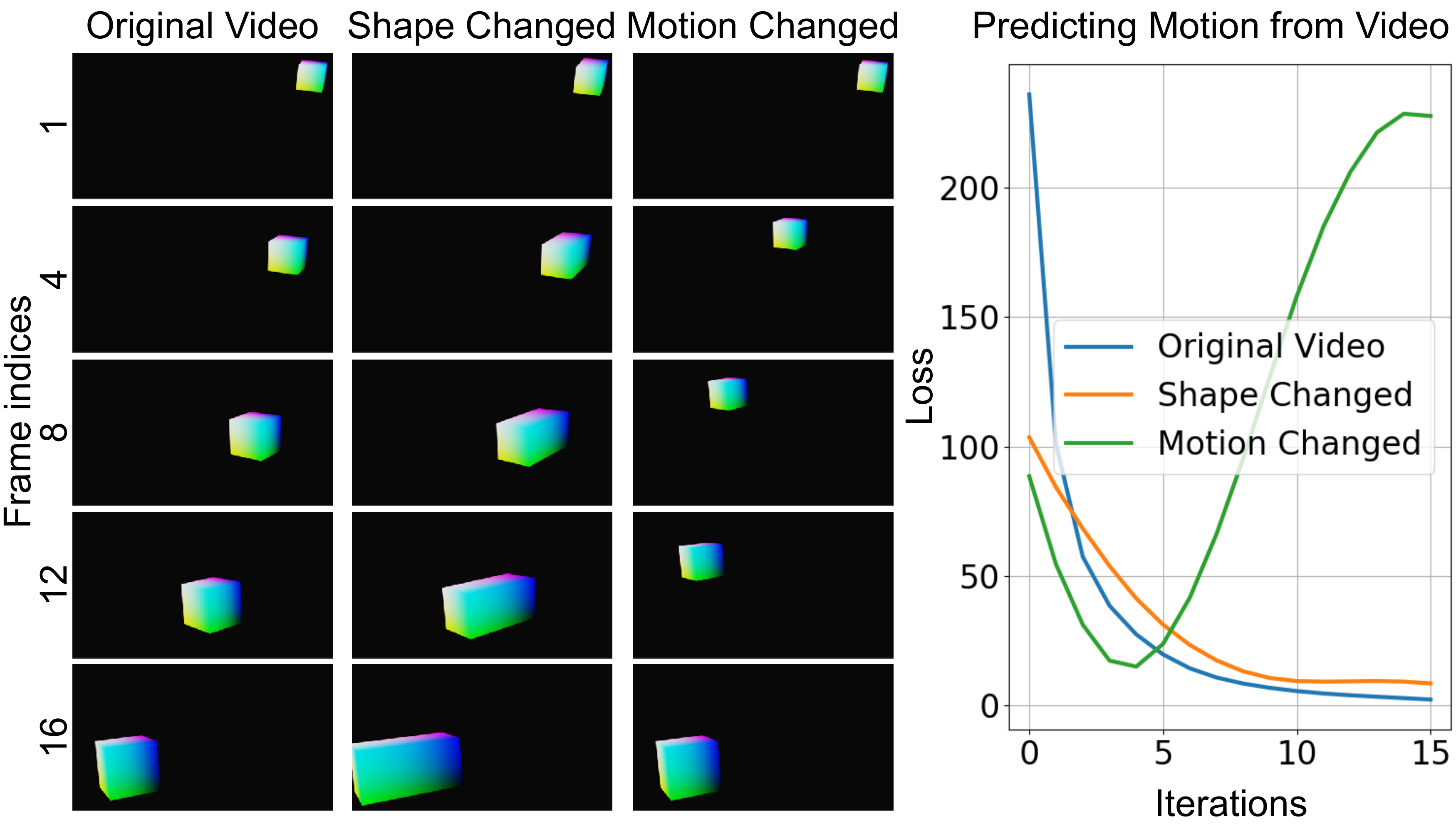}
    \caption{\textbf{Toy experiment on learning shape-invariant motion.} We trained our \textsc{MotionGuide} module on the original video and tested it on videos with 1) shape changes, and 2) motion changes. We show the frames for each video to the left. From the graph at the right, it may be observed that the \textsc{MotionGuide} module is invariant to shape change but sensitive to motion change.}
    \label{fig:toy}
\end{figure}

\begin{figure*}
    \centering
\includegraphics[width=0.93\linewidth]{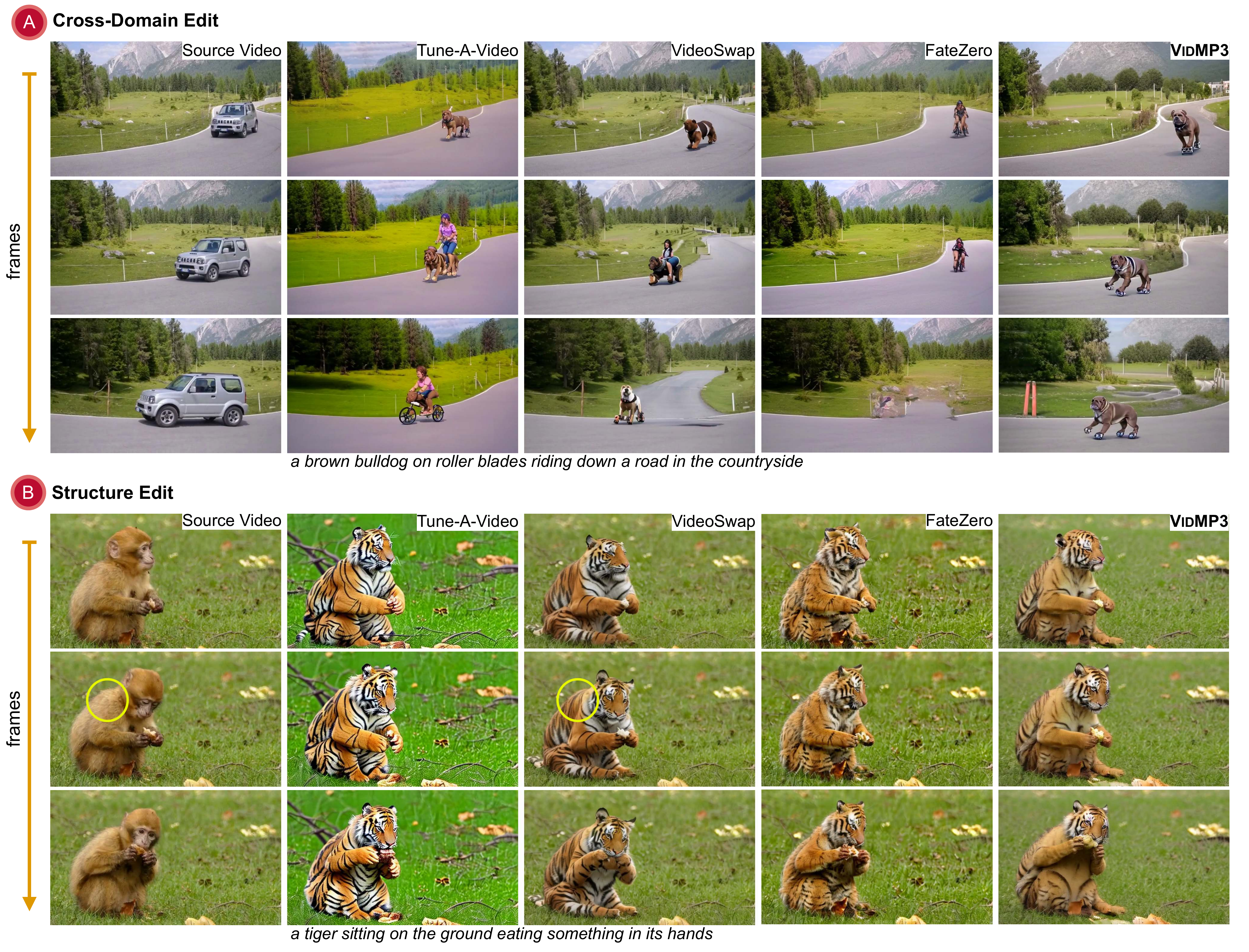}
    \caption{\textbf{Comparison with prior art on motion-preserved video editing.} We consider the challenging cases of \mcircle{A} \textbf{Cross-Domain Edit} -- ``silver jeep" $\xrightarrow{}$ ``bulldog on roller blades", and \mcircle{B} \textbf{Structure Edit} -- ``monkey"$\xrightarrow{}$ ``tiger". It may be observed that in the case of cross-domain editing, all baselines suffer from severe temporal inconsistencies of the subject. For the case of structure editing, Tune-A-Video produces a highly saturated video with the head pose not correctly following the pose of the input video. Similarly, FateZero also models incorrect head pose (see second row of \mcircle{B}). For VideoSwap we notice that the tiger has a similar humped shape like the monkey (notice the yellow circled areas), due to the keypoint correspondences being very sparse and spatially constrained signal. The sparsity of this signal results in the orientation of the face being inaccurate, resulting in a wrong head pose of the tiger in the middle row. By comparison, \textsc{VidMP3} generates temporally consistent results following the input pose while making necessary changes faithful to the new concept.\vspace{-2mm}}
    \label{fig:compare}
\end{figure*}

\begin{figure}
    \centering
\includegraphics[width=1.0\linewidth]{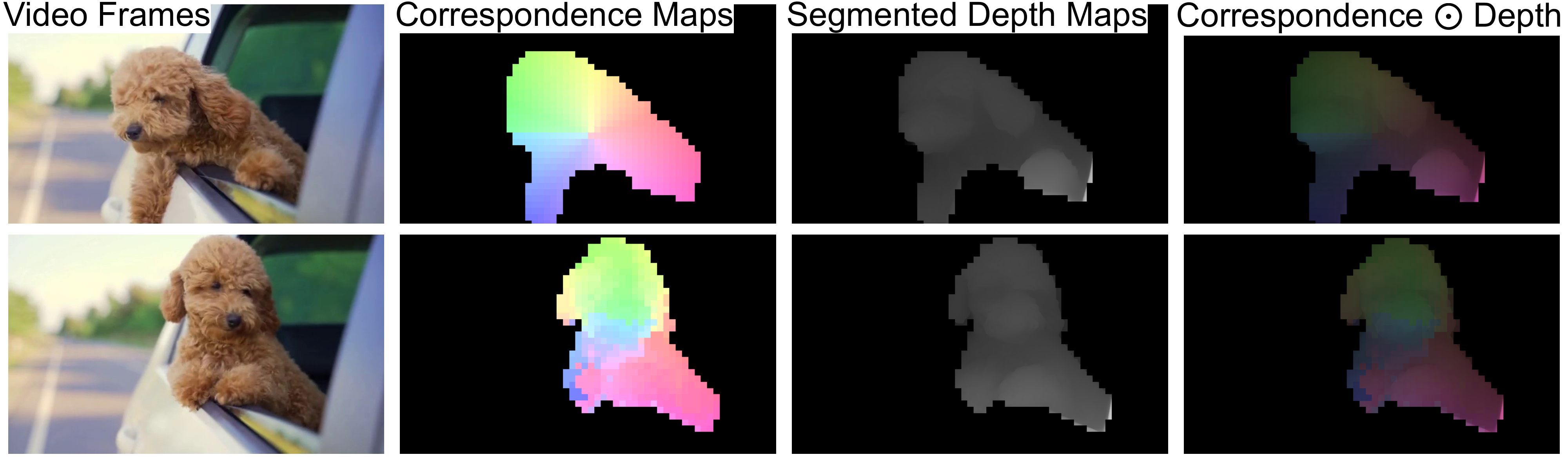}
    \caption{\textbf{Visualization of correspondence and depth maps.} For two frames of the video of ``a dog looking out the window of a car", we show the corresponding correspondence and depth maps we obtain from off-the-shelf models. The depth segmented using the correspondence map is multiplied with the correspondence map (right-most column) and provided as input to \textsc{MotionGuide}.}
    \label{fig:depthcorr}
\end{figure}

\begin{figure*}
    \centering
\includegraphics[width=0.9\linewidth]{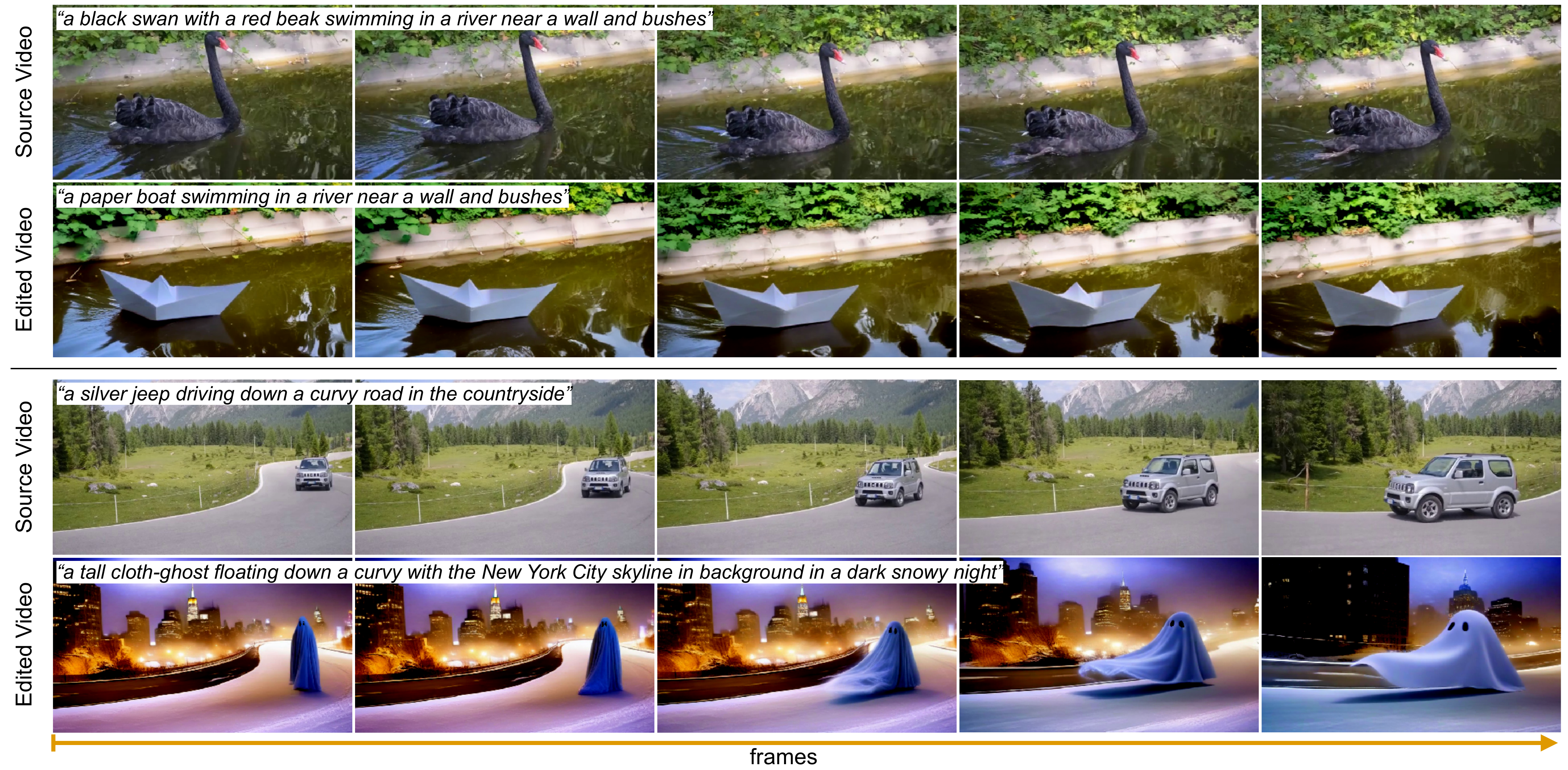}
    \caption{\textbf{Scaling \textsc{VidMP3} to SDXL.} As a novel initiative, we scaled up the T2V model to utilize SDXL as the foundation model. We show that we can model more diverse concepts using this setup, owing to the stronger generation capabilities of SDXL.}
    \label{fig:sdxl}
\end{figure*}

\section{Method}
The motion of any object can be represented as a combination of pose and position in 3D space. Given a video $\mathbf{X}_N = [x_1, x_2 \dots x_N]$ of $N$ frames, we wish to learn only the motion of the subject in the video. We want to build a \textbf{generalized representation of motion} using the 3D pose and position of an object. This representation enables us to swap objects with significantly different shapes or semantics. We hypothesize that motion can be extracted only using the dense correspondences within frames $\mathbf{C}_N$ and the depth maps per frame $\mathbf{D}_N$, without using the frames of the video $\mathbf{X}_N$. $\mathbf{C}_N$ is useful for representing the 2D position and pose of the object, while $\mathbf{D}_N$ represents the 3D position. First, we present a proof of concept, whereby we introduce a \textsc{MotionGuide} module to learn motion using $\mathbf{C}_N$ and $\mathbf{D}_N$, and show that the learned representation of the module is invariant to shape changes but sensitive to changes in motion. Next, we formally describe how the representations of this \textsc{MotionGuide} module can be injected into a diffusion model to edit videos.

\subsection{Representing motion with pose and position}
We design a \textsc{MotionGuide} module $\phi_m$ that takes as input dense correspondence maps $\mathbf{C}_N$ and depth maps $\mathbf{D}_N$ of the subject of interest in a video. We present the design of this lightweight module in the Appendix. Essentially, the module processes $\mathbf{C}_N$ $\odot$ $\mathbf{D}_N$ with convolution layers, then concatenates a positional encoding $\mathbf{P}$ to each frame. After another convolution, we average pool in the spatial dimensions and divide by $\alpha$ to form a single-dimensional vector for each frame $\mathbf{M}_{N,d}$ , where $\alpha$ is the ratio of pixels occupied by the object in the frame to the total number of pixels in the frame. This is then processed by a final linear layer. The pooling is crucial to our method as it prevents shape and size leakage. The positional encoding $\mathbf{P}$ provides information on the 2D location of the values in $\mathbf{C}_N$ $\odot$ $\mathbf{D}_N$, making the representation sensitive to the average 2D position, even after spatial pooling.

\subsection{Toy experiment}
For proof of concept, we designed a toy experiment where the \textsc{MotionGuide} module $\phi_m$ was attached with a final linear layer to predict the 3D trajectory and rotations of an object. We rendered a video of a cube following a specific trajectory and rotations $\mathbf{T}_{N,6}$. The 6 values correspond to positions in xyz and rotations in xyz. The cube is rendered with different gradient colors on its faces to mimic correspondence maps. We treated the rendered frames of the cube as correspondence maps $\mathbf{C}_N$ and found depth maps of each frame, denoted as $\mathbf{D}_N$. We trained $\phi_m$ on this single video of the cube to predict $\mathbf{\hat{T}}_{N,6}$ by optimizing: 

\begin{equation}
\min_{\phi_m} \left\| \mathbf{T}_{N,6} - \phi_m(\mathbf{C}_N, \mathbf{D}_N) \right\|^2 .
\end{equation}

Given this trained \textsc{MotionGuide} module $\phi_m$, we used it to infer on 1) ${\mathbf{C}^1}_N$, ${\mathbf{D}^1}_N$ of a positive sample where the shape of the cube was changed, but followed the same motion, and 2) ${\mathbf{C}^2}_N$, ${\mathbf{D}^2}_N$ of a negative sample where the original cube followed a different motion. We present frames of the original video, and the test videos along with prediction loss in 
Fig.~\ref{fig:toy}. It may be observed that the training loss and loss of the positive sample follow a similar reducing trend, while that of the negative sample diverges. This shows 1) that motion can be predicted reasonably using correspondence and depth maps, 2) the learned representation is invariant to shape change, and 3) the learned representation is sensitive to motion changes.

\subsection{\textbf{\textsc{VidMP3}}}
\textsc{VidMP3}, depicted in Fig.~\ref{fig:vidmp3}, utilizes the \textsc{MotionGuide} module formulated in the previous section to learn motion from a source video $\mathbf{X}_N$, to generate a new video having the same motion. We fine-tuned our model on the single source video $\mathbf{X}_N$. We followed the paradigm of Tune-A-Video~\cite{wu2023tune}, where motion modules are inserted into a pre-trained T2I diffusion model. The motion module consists of temporal self-attention layers which are computed as:
\begin{align}
    \text{Attention}(\mathbf{Q}, \mathbf{K}, \mathbf{V}) = \text{softmax}\left(\frac{\mathbf{Q} \mathbf{K}^\top}{\sqrt{d}}\right) \mathbf{V}, \\
    \mathbf{Q} = \mathbf{W^Q} \mathbf{z}_{i,j} , \quad \mathbf{K} = \mathbf{W^K} \mathbf{z}_{i,j} , \quad \mathbf{V} = \mathbf{W^V} \mathbf{z}_{i,j} ,
\end{align}
where $\mathbf{z}_{i,j}$ is the latent representation of the video at a spatial location $(i,j)$ before the temporal self-attention. We inject the output of our \textsc{MotionGuide} module into the values of the temporal self-attentions such that:
\begin{equation}
    \mathbf{V} = \mathbf{W^V} (\mathbf{z}_{i,j} + \lambda \phi_m(\mathbf{C}_N, \mathbf{D}_N)),
\end{equation}
where $\lambda$ is a weighting factor. We chose to inject the external features into the values, to add extra context to the locations the self-attention focuses on. We used the pre-trained weights of the motion module from AnimateDiff~\cite{guo2023animatediff}.

We updated the spatial self-attention to the sparse causal variant of Tune-A-Video, where for a specific frame the attention is calculated using the first and previous frame of the video. Unlike Tune-A-Video which suffers from severe shape leakage because of over-fitting the full motion modules on the source video, we chose to keep the motion module frozen and inject motion only using the external adapter \textsc{MotionGuide} module. This enables us to learn a representation space of pure motion disentangled from appearance. We trained this modified network by optimizing:
\begin{equation}
     \min_{\phi_m, \phi_u} \mathbb{E}_{z_0, t, \epsilon}\left[ \|\epsilon - \epsilon_\theta(z_t; t, y, \phi_m(\mathbf{C}_N, \mathbf{D}_N)\|^2 \right],
\end{equation}
where $t$ represents the time-step, $z_t$ the latents diffused at time $t$, $y$ the prompt for the source video, and $\epsilon_\theta$ represents the denoising diffusion model. We optimized only over $\phi_m$ and $\phi_u$. $\phi_u$ represents other trainable parameters, namely $\mathbf{W^Q}$ of the spatial self- and cross-attention layers, and $\mathbf{W^V}$ of the motion modules. Finally, after training, we used the inverted latents of the source video to sample a new video with an edited prompt, while using $\mathbf{C}_N$ and $\mathbf{D}_N$ of the source video. We show that this simple formulation is highly robust and quite general, enabling us to generate subjects that are significantly different in shape and semantics as compared to the subject in the original video.

\section{Experiments} 
\label{sec:experiments}
\paragraph{Datasets.}
We used the same set of 30 videos provided by VideoSwap which were selected from Shutterstock and DAVIS~\cite{pont20172017}. The videos are divided into three categories -- human, animal, and object -- where each category comprises of 10 videos. For each source video we used three predefined concepts and three customized concepts, resulting in a total of 180 edited videos. Unlike VideoSwap, our customized concepts involve significant semantic changes.

\paragraph{Implementation Details.} We used Stable Diffusion 1.5 as the foundation model for baseline comparisons and also extended our method to use SDXL for generating more diverse concepts. For the SD-1.5 architecture, we primarily use Chilloutmix~\cite{chilloutmix} pre-trained weights, except for 1) style editing where we used the original SD-1.5 weights, or 2) personalized editing tasks. We used the pre-trained motion modules of AnimateDiff~\cite{guo2023animatediff} for the temporal self-attention layers. We uniformly sampled frames at a sampling rate of 4 at their original resolution from the input video to finetune the models. All experiments were conducted on Nvidia A100 (40GB) and H100 GPUs. We used Adam with a learning rate of $5e^{-4}$ when optimizing the fine-tuning stage over 100 iterations. We set the \textsc{MotionGuide} weighting factor $\lambda$ to a value of $0.1$ for videos with higher ranges of motion and $0.05$ for videos with lower ranges of motion. The weights of the final linear layer of the \textsc{MotionGuide} module are zero-initialized when training so that the output of the \textsc{MotionGuide} module is zero for the first iteration. We also disabled the bias of the convolution layers of the \textsc{MotionGuide}, since we are overfitting on one video without the need to have any regularization.

To compute spatial correspondence maps, we used the implementation of SD-Dino~\cite{zhang2024tale}, which utlizes the internal deep features of Dino~\cite{caron2021emerging} and Stable Diffusion~\cite{rombach2022high} for this task. For classes not present in the COCO dataset e.g. ``monkey", we used the off-the-shelf figure-ground segmentation tool RMBG-1.4~\cite{RMBG-1.4}. We found correspondence maps for each frame using the first frame as reference. Depth maps were found using DepthAnythingV2~\cite{yang2024depth}, which are then segmented to only contain the subject aided by the obtained correspondence maps. Finally we multiplied the correspondence and segmented depth maps to form the input to \textsc{MotionGuide}. We show examples of the computed correspondence and depth maps for a video in Fig.~\ref{fig:depthcorr}.

\paragraph{Baselines.}

We qualitatively and quantitatively compared our model to Tune-A-Video~\cite{wu2023tune}, VideoSwap~\cite{gu2024videoswap}, and FateZero~\cite{qi2023fatezero}.
We found these baselines to be the most relevant ones delivering the strongest results for motion-preserved editing tasks using a single video for training.\footnote{CAMEL~\cite{zhang2024camel} is a related work but does not provide sufficient results, and omits dependencies required to run their code in their repository.} We show in the Appendix that first-frame editing methods like AnyV2V struggle to capture considerable levels of motion and are highly dependent on the quality of the first frame generated by their image editing method.

\section{Results}

Here we showcase some of the various capabilities of \textsc{VidMP3}, comparison to baselines, adaptability of our model to various video editing tasks, scaling to SDXL, ablations over the components of our method, and discuss implementation choices.

\paragraph{Cross-domain Edit.}
The most important contribution of  \textsc{VidMP3} lies in the challenging case of Cross-domain Editing, where previous methods suffer. In this case, we show that the subject in the source video can be swapped with a semantically different subject in the edited video, while correctly preserving motion. In Fig.~\ref{fig:compare} we show the instance ``silver jeep" $\rightarrow$ ``bulldog on roller blades," where \textsc{VidMP3} can generate a video where the motion is preserved and the subject is temporally consistent. We attribute these results to the external strong motion signal we inject, which allows the model to understand a general sense of position and pose. We present additional results in the Appendix.

\paragraph{Structure Edit.}
Previous methods have shown good performance for the case of structure editing, while keeping the edited subject in the same domain, e.g., ``silver jeep" $\rightarrow$ ``Porsche." This case is much simpler as compared to cross-domain editing, due to the internal semantic understanding of the diffusion model. We show the case of ``monkey" $\rightarrow$ ``tiger" in Fig.~\ref{fig:compare}, where the edited tiger generated by \textsc{VidMP3} follows the exact same head and hand motion as the monkey, allowing freedom for the different body shapes of the tiger as compared to the monkey.  We present additional results for structure editing in the Appendix.

\paragraph{Comparison to baselines.}
For the two previously described cases of \textbf{Cross-Domain Edit} and \textbf{Structure Edit}, we compared to the previous methods, Tune-A-Video, VideoSwap and FateZero. For fair comparison, we initialized all baselines with the same pre-trained T2I weights~\cite{chilloutmix} as ours. Tune-A-Video and FateZero don't explicitly provide any external guidance to the model, which lead to high temporal inconsistencies in the case of Cross-Domain Editing, where the pre-trained T2I model is not confident in its outputs owing to semantic changes of the object to be edited with respect to the source object. On the other hand, VideoSwap uses explict keypoint correspondences and guides the model to change the object, but it fails when the semantic meanings do not remain relevant (e.g.: ``silver jeep" $\rightarrow$ ``brown bulldog"). VideoSwap requires human effort in marking the positions of 2D keypoints that should be tracked in the video. It also involves significant time and human effort to manually edit the position of the keypoints for the target video when there are significant shape changes. Tune-A-Video generates saturated videos on both Cross-Domain and Structure Editing, possibly due to overfitting the entire motion module on the source video. This is not true for either VideoSwap or \textsc{VidMP3}, as all or most parts of the motion module are kept frozen while learning an external adapter that has a fixed input. For the case of Structure Editing, it may be observed that VideoSwap generated the tiger to be in a bent posture like the monkey, because fitting to the keypoint correspondence signal was too constrictive. None of the baselines were able to follow the head pose of the monkey accurately as can be observed especially in the second and third row of the generated videos of all baselines in Fig.~\ref{fig:compare} \mcircle{B}.

By contrast, \textsc{VidMP3} generates temporally consistent videos in both cases while preserving motion from the source video. This is achieved by computing a pose and position representative value for each frame, using dense correspondence and depth maps to learn generalized representation of motion. During inference, these representations help guide motion in the generated videos, and allowing the text-to-image model more room to explore appearances. 


\paragraph{Adaptability of \textsc{VidMP3}.}
Since \textsc{VidMP3} is based on an existing T2I model, it can be effectively applied to tasks other than subject swapping. We show results of using \textsc{VidMP3} for 1) background change, 2) style change, and 3) personalization. For personalization, we attempted both per-subject personalization, as well as using pre-trained T2I models that are personalized on more general concepts, such as Anything-v4.0~\cite{Anything-v4.0}. We refer the readers to the Appendix for qualitative results of these tasks.

\paragraph{Scaling \textsc{VidMP3} to SDXL.}
We also studied scaling to StableDiffusion-XL which is a stronger T2I model that is able to represent more diverse concepts. We use AnimateDiff's SDXL motion module, and found that it less effectively models motion than the motion module of the SD-1.5 version. Thus, we identify the specific parameters of the motion module that contribute to shape leakage and kept them frozen. More specifically, we found that the feed-forward layers of temporal self-attention blocks contribute to the highest leakage. We trained the other parameters namely, $\mathbf{W^Q}$, $\mathbf{W^K}$, and $\mathbf{W^V}$ and projection matrices of the temporal self-attention modules, in addition to the parameters that we kept trainable in the SD-1.5 version. This enabled our model to better learn motion while still avoiding leakage. We present results of using SDXL within \textsc{VidMP3} in Fig.~\ref{fig:sdxl}. We show generated concepts that we were not able to represent consistently using SD-1.5 \textsc{VidMP3}, such as a ``paper boat" and a ``cloth-ghost". We provide additional results generated by \textsc{VidMP3} SDXL in the Appendix.

\begin{figure*}
    \centering
\includegraphics[width=0.9\linewidth]{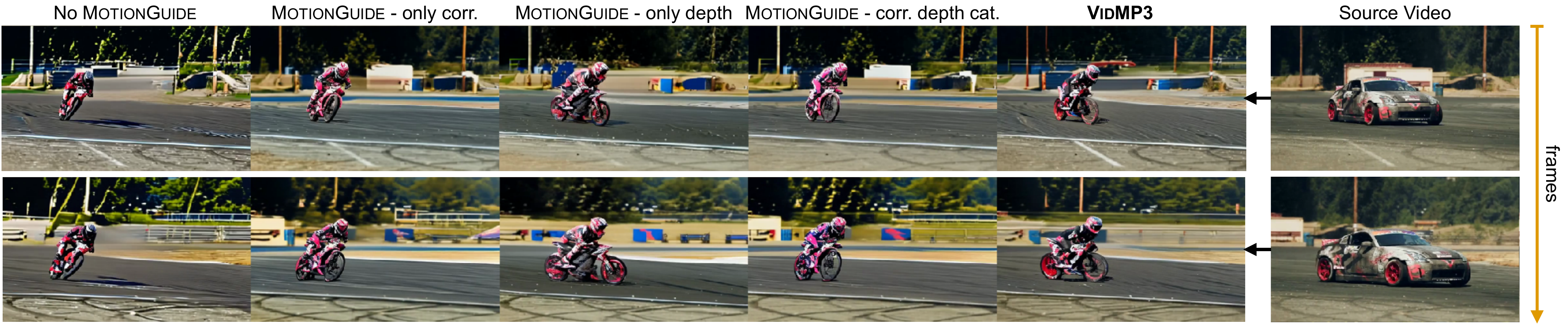}
    \caption{\textbf{Ablation over various components of our method.} For the case of ``car" $\rightarrow$ ``bike", we see that all other implementations including disabling \textsc{MotionGuide}, providing different inputs to \textsc{MotionGuide} such as only correspondence maps, only depth maps or concatenation of depth and correspondence maps, results in incorrect and lower range of motion. \textsc{VidMP3} uses \textsc{MotionGuide} with multiplied correspondence and depth maps as input and imitates the motion the subject in the source video correctly.\vspace{-2mm}}
    \label{fig:ablation}
\end{figure*}

\paragraph{Ablations.} 
We conducted ablations over various components of our method and implementation choices. Fig.~\ref{fig:ablation} depicts the effect of using only correspondence maps, only depth maps, or concatenated depth and correspondence maps as input to \textsc{MotionGuide}. We also show the effect of disabling the \textsc{MotionGuide}. For all these cases, we find that the motion is modeled incorrectly, with a much subdued range and incorrect orientations per frame.



\begin{figure}
    \centering
    \includegraphics[width=0.95\linewidth]{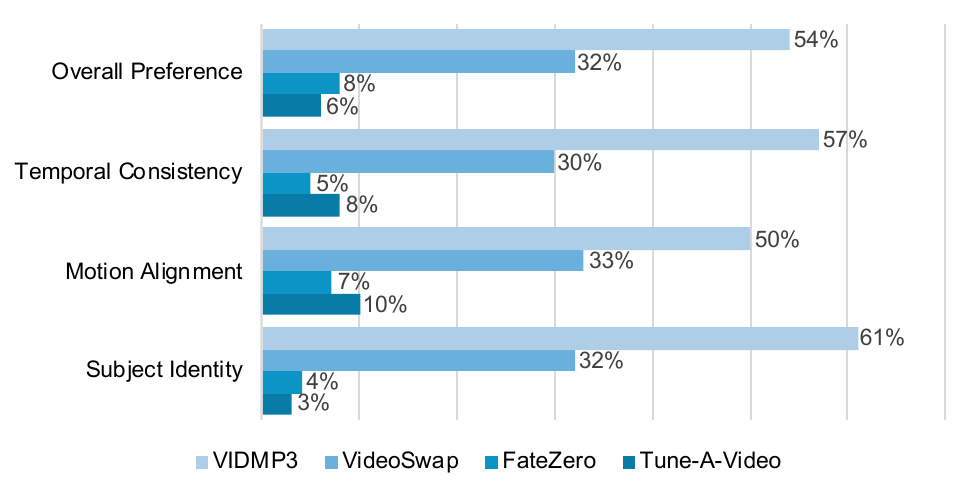}
    \caption{\textbf{Human opinions} on 1) Subject Identity, 2) Motion Alignment, 3) Temporal Consistency, and 4) Overall Preference, averaged over 10 participants and 180 edited videos.\vspace{-2mm}}
    \label{fig:humanStudy}
\end{figure}

\paragraph{Evaluation.}
We quantitatively compared our method against previous SOTA models using both automatic and human evaluations. We provide a detailed discussion of the evaluation settings in the Appendix. We conducted a voluntary, controlled laboratory human study to gather opinions expressive of 1) Subject Identity, 2) Motion Alignment, 3) Temporal Consistency, and 4) Overall Preference for video subject swapping. The results of this evaluation, shown in Fig.~\ref{fig:humanStudy}, indicate a clear preference for our method.

\paragraph{Time Cost Analysis}
We recorded the time required to run each component of \textsc{VidMP3} to edit a 16 frame video clip on an Nvidia A100 GPU. This includes 1) \textit{Preprocessing}:  which involves computing the correspondence and depth maps. The correspondence map computation required approximately 4s per frame, or 64 seconds over 16 frames. The depth map computation required approximately 2s per frame, or 32 seconds over 16 frames. The preprocessing step used about 100 seconds overall. 2) \textit{Training}:  where the \textsc{MotionGuide} module was trained over 100 iterations which expended about 3 minutes of compute time. And, lastly 3) \textit{Editing}:  when we generated the edited video using inverted noise from the source video. The DDIM inversion process of 50 steps required about 30 seconds. The backward process to generate the edited video consumed about 30 seconds as well, resulting in a total of 60 seconds. Overall, the complete process, from preprocessing to generating the final edited video required about 6-7 minutes. 

\vspace{-1mm}
\section{Limitations and Discussions}
There can be multiple choices of features that could represent the pose and position of subjects, and that can be injected externally to a diffusion model to guide motion, e.g. injecting Diffusion Correspondence (DIFT~\cite{tang2023emergent}) features. However, our choice of 2D correspondence and depth maps is highly efficient since it only requires three channels of input, and is also a cleaner signal of explicit motion without any leakage of extra information. While \textsc{VidMP3} can generate subjects with significant structural and semantic differences relative to the source video, we cannot explicitly control the size of the subject. For example, in the case of ``black swan" $\rightarrow$ ``paper boat" in Fig.~\ref{fig:sdxl}, observe that the generated paper boat is large and of a similar size as the swan. Additionaly, our method is dependant on the quality of correspondence and depth maps obtained. However, for all of our evaluation videos, we find that the off-the-shelf methods for obtaining these maps perform well. Scaling video editing to multiple subjects has been studied in previous work, but has not been explored here. For such scenarios, one approach would be to generate separate correspondence maps for each of the various subjects of interest, and inject each using separate \textsc{MotionGuide} modules. We leave this direction of research for future work.



\vspace{-2mm}
\section{Conclusion}
We presented \textsc{VidMP3}, a novel video editing technique based on T2I models, which utilizes pose and position priors to generate motion-preserved videos based on a source video. We introduce the \textsc{MotionGuide} module, which learns generalized motion representations from spatial correspondence and depth maps. These representations are injected into the temporal self-attention layers of a T2V model initialized from a T2I model, thus forming \textsc{VidMP3}. We evaluated \textsc{VidMP3} on challenging video editing tasks: 1) Cross-Domain Editing, and 2) Structure Editing. We observed that \textsc{VidMP3} can generate objects with significant structural and semantic changes relative to the subject in the source video, while maintaining temporal consistency. We show qualitatively and quantitatively that our method improves over previous strong baselines on the task of motion-preserved video editing. Additionally, we scaled our method to use SDXL as the base T2I model, which is a novel effort in the area of video editing. We explored the adaptibility of our method on various video editing tasks, including personalized editing, background editing, and style editing. Despite its potential to enhance creative workflows, motion-preserved video editing without rigid structural constraints remains a relatively under-explored domain. \textsc{VidMP3} addresses this gap by introducing a novel approach that maintains temporal coherence while allowing flexible content modification, laying the groundwork for future research and developments in this area.


\section*{Acknowledgment}
The authors thank the Texas Advanced Computing Center (TACC) for providing computational resources that contributed to this research, and the NTIA and NSF for supporting this work. 

{
    \small
    \bibliographystyle{ieeenat_fullname}
    \bibliography{main}
}

\clearpage
\setcounter{page}{1}
\maketitlesupplementary

\section{Adaptability of \textsc{VidMP3}}

\begin{figure*}[!hb]
    \centering
    \includegraphics[width=0.94\linewidth]{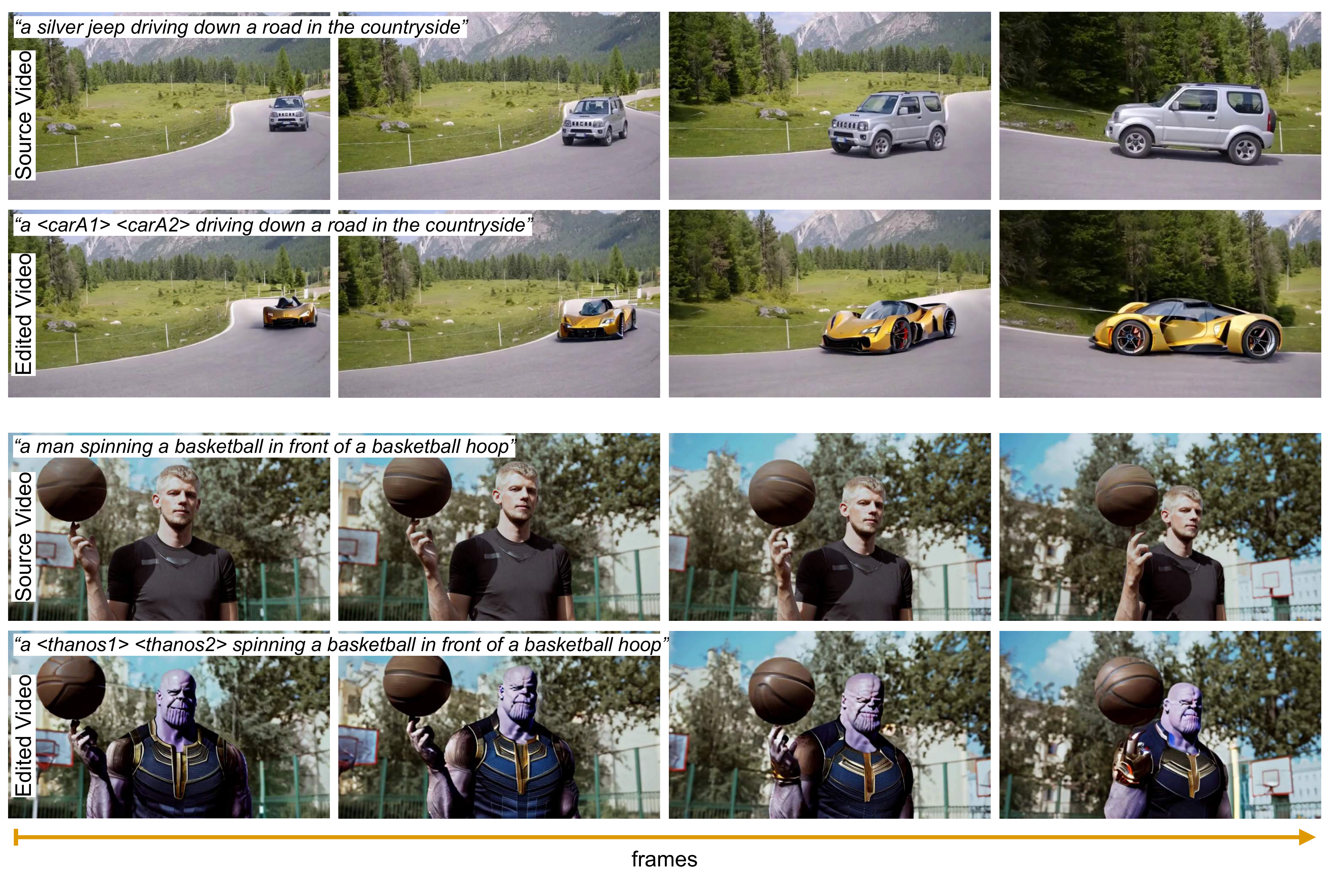}
    \caption{\textbf{Personalization.} Using pre-trained ED-LoRA concepts during inference, we generated the illustrated frames featuring personalized subjects: a concept car (top) and the character Thanos (bottom).}
    \label{fig:personlization_results}
\end{figure*}

\textbf{Personalization.} \textsc{VidMP3} supports customized or personalized concepts through additive methods such as ED-LoRA~\cite{gu2024mix}. In Fig.~\ref{fig:personlization_results}, we present edited videos generated using \textsc{VidMP3} with pre-trained customizations provided by ~\cite{gu2024videoswap}. During model optimization on the source video, the LoRA layers were not attached; they were utilized only during inference on the saved model. The degree of customization can be adjusted using the LoRA blend weight parameter.

\begin{figure*}[t]
    \centering
    \includegraphics[width=0.95\linewidth]{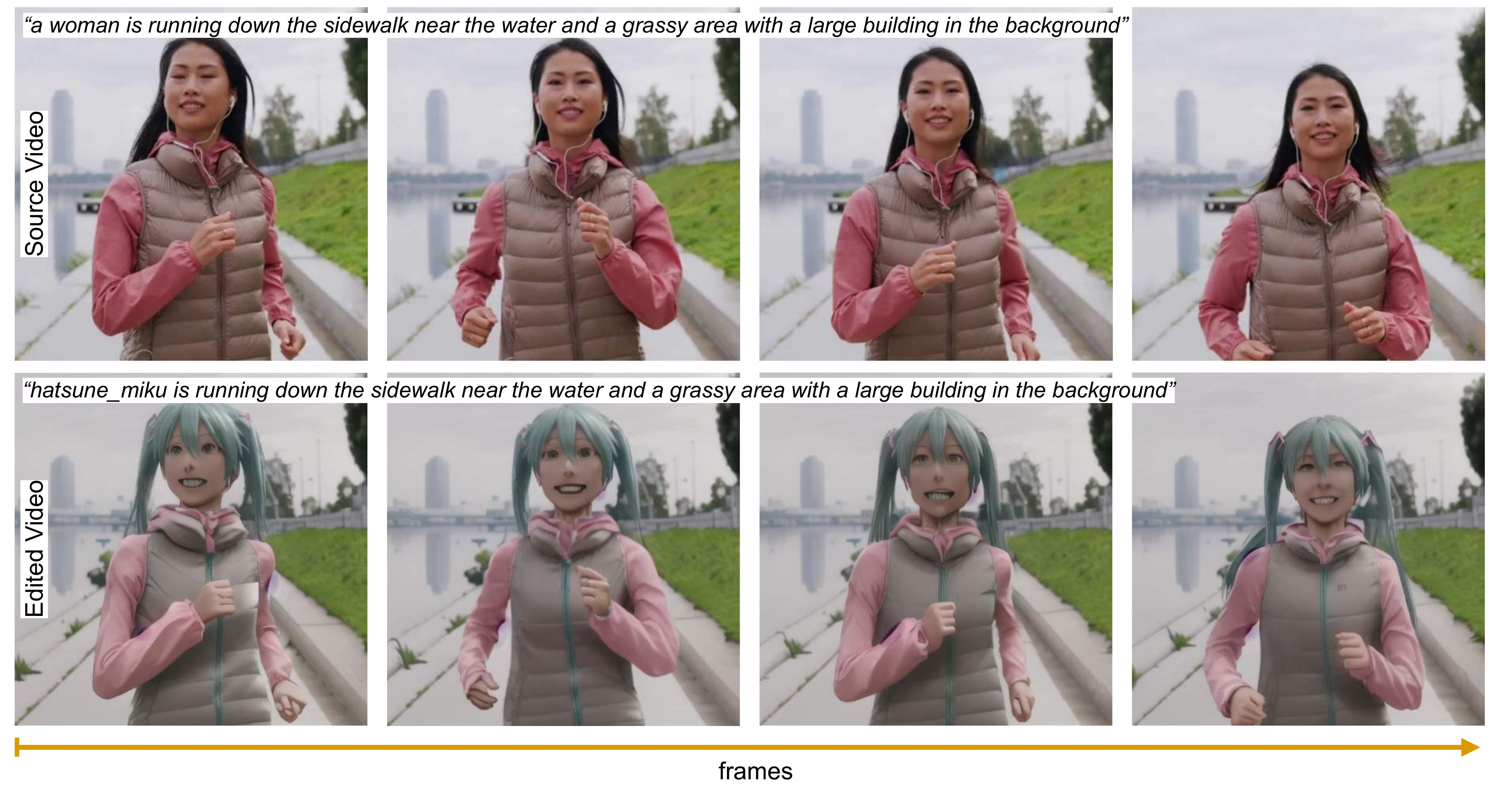}
    \caption{\textbf{Theme Personalization} Edited video frames rendered in anime style using Anything-v4.0 as the foundation model in \textsc{VidMP3}.}
    \vspace{-10pt}
    \label{fig:personlization_anythingv}
\end{figure*}

In Fig.~\ref{fig:personlization_anythingv}, we present editing results generated using \textsc{VidMP3} with the Anything-v4.0 personalized model as the foundation model. This model specializes in producing anime-style images.

\begin{figure*}[b]
    \centering
    \includegraphics[width=0.95\linewidth]{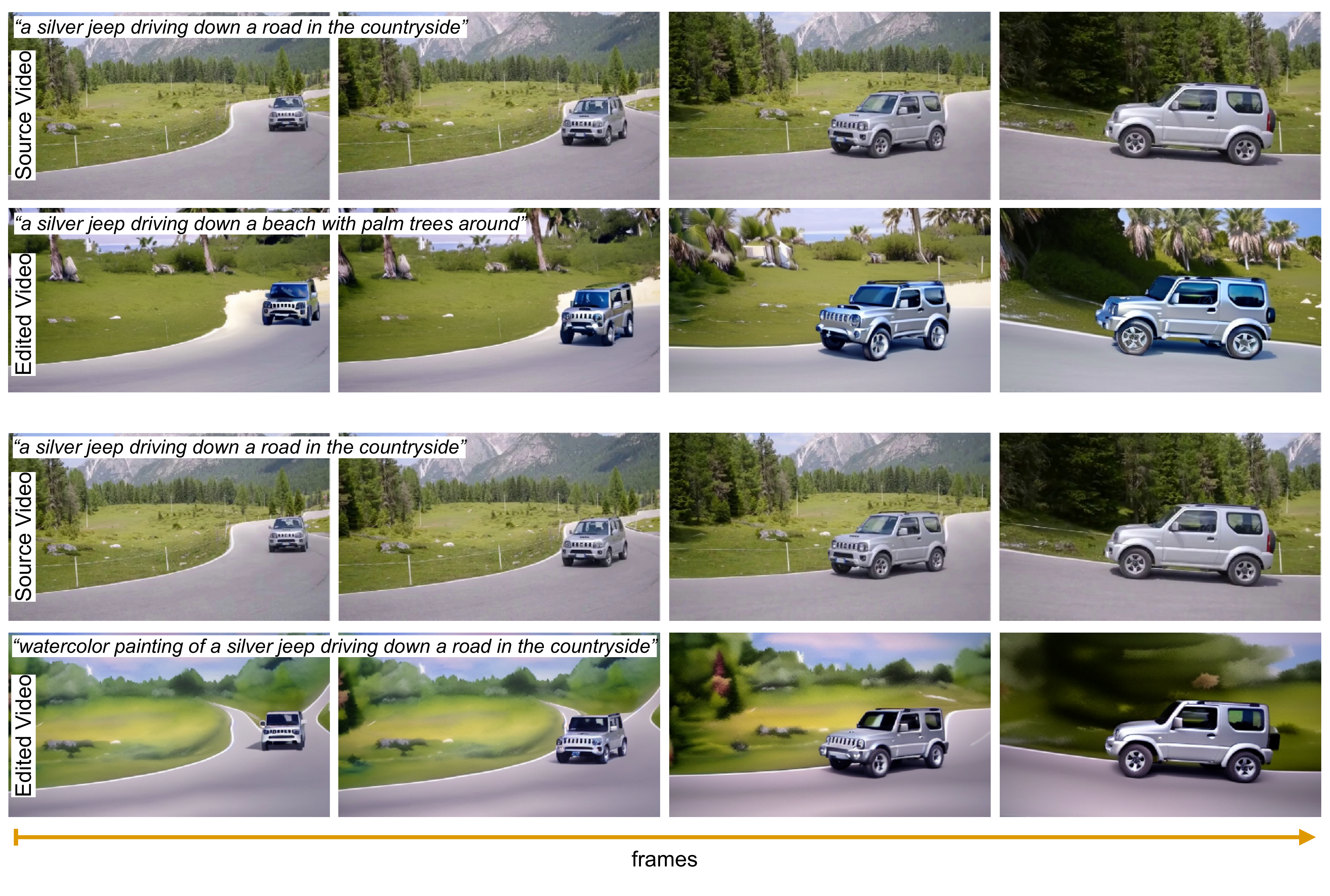}
    \caption{\textbf{Background and Style Edit.} Results of background modification (top) and style modification (bottom) using SD-v1.5 as the foundation model in \textsc{VidMP3}.}
    \label{fig:BGandStyleEdit}
\end{figure*}

\begin{figure*}[b]
    \centering
    \includegraphics[width=0.94\linewidth]{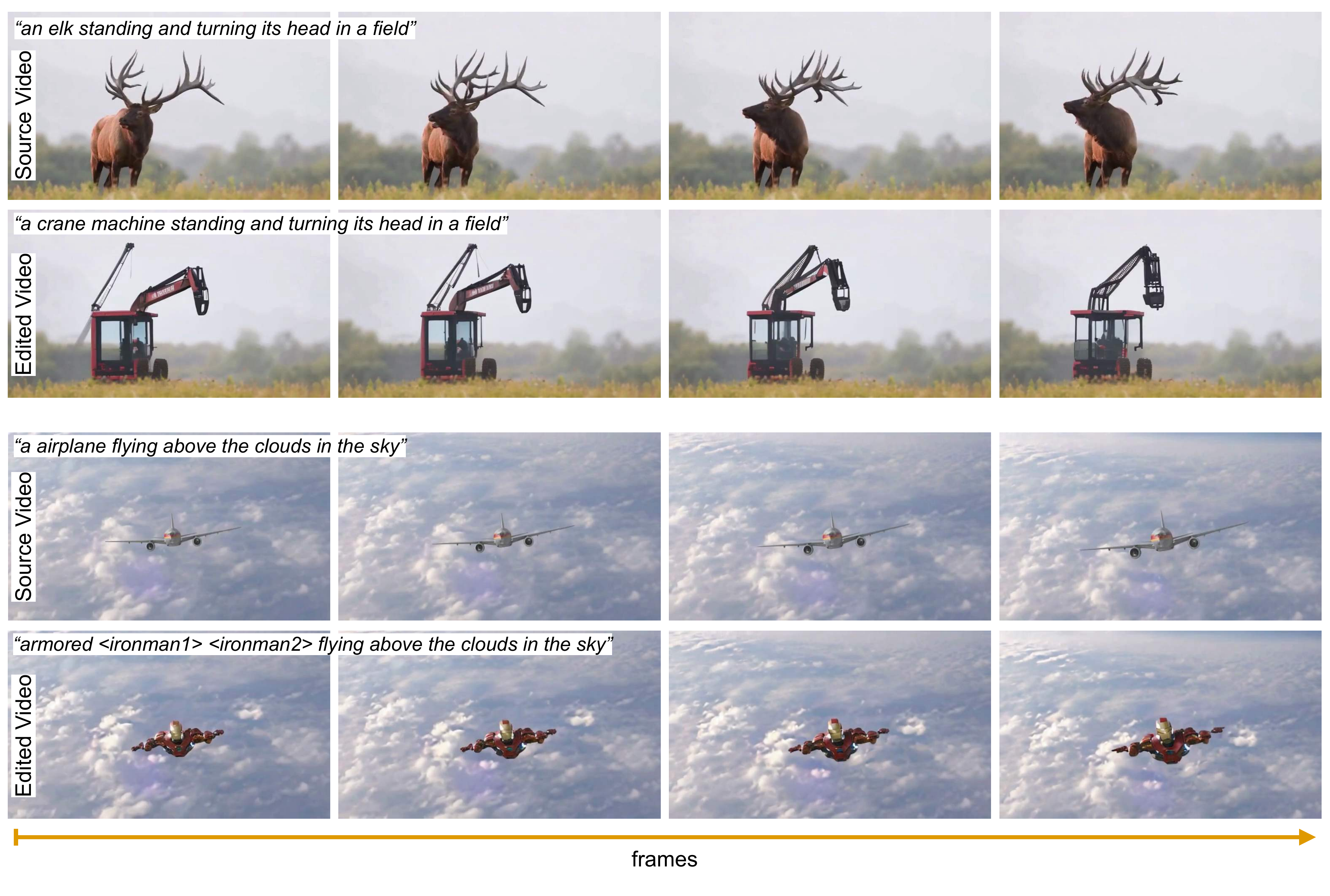}
    \caption{\textbf{Cross-Domain Edit.} Examples of cross-domain edits where an animate object is replaced with an inanimate object (top: ``elk" $\rightarrow$ ``crane machine") and an inanimate object is replaced with an animate object (bottom: ``airplane" $\rightarrow$ ``Ironman").}
    \label{fig:crossdomain_appx}
\end{figure*}

\textbf{Background and Style Editing}
We also explore background and style edits, with the results shown in Fig.~\ref{fig:BGandStyleEdit}, illustrating background edit in the second row and style edits in the fourth row.

\begin{figure*}[t]
    \centering
    \includegraphics[width=0.94\linewidth]{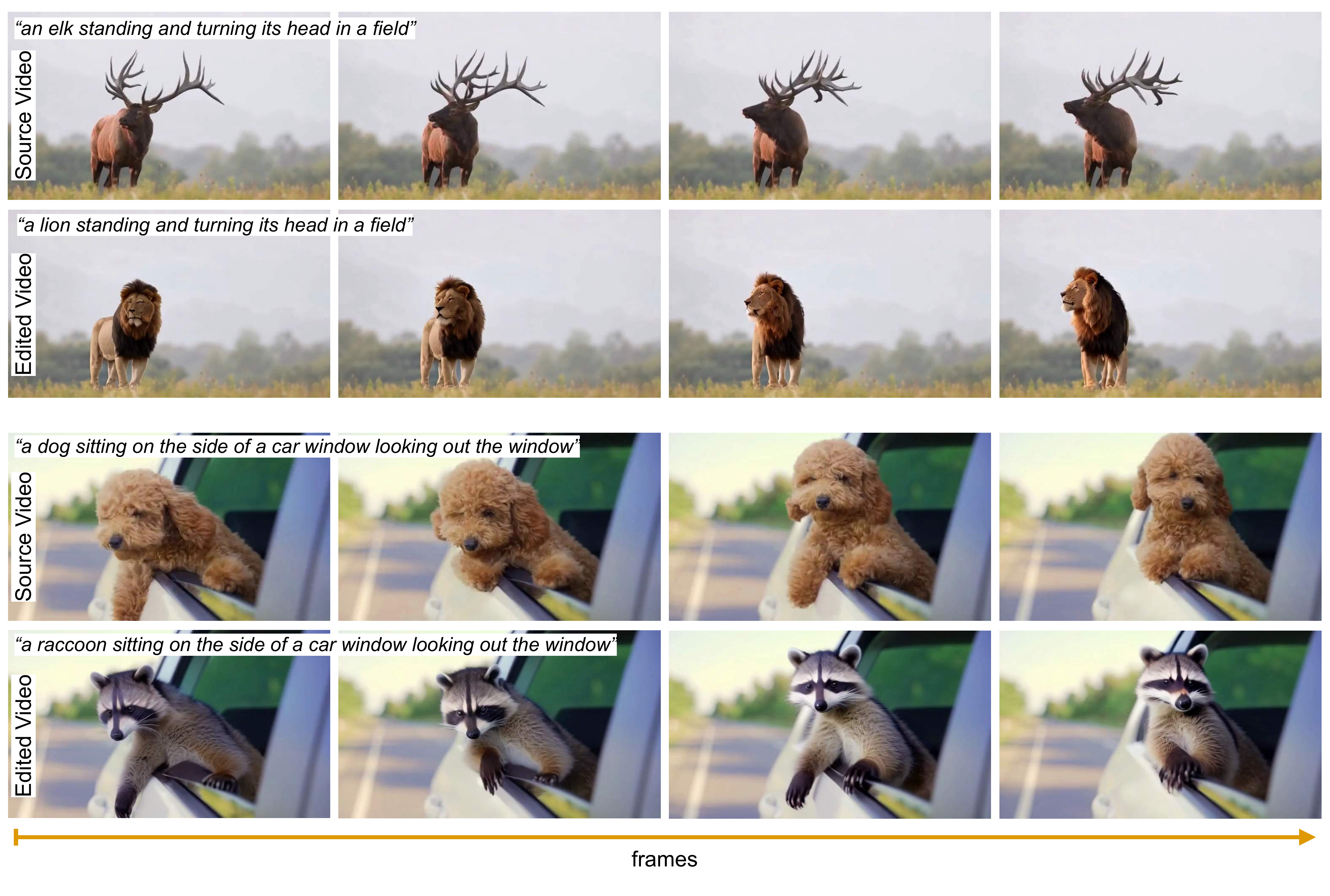}
    \caption{\textbf{Structure Edit.} Examples of structural editing, while keeping the edited subject in the same domain.}
    \label{fig:structure_appx}
\end{figure*}

\textbf{Latent Blending.} Our proposed method is robust enough to incorporate plug-and-play features such as latent blending as used in VideoSwap. Latent blending facilitates subject swapping while preserving the background region in the edited video to remain identical to the source video. The core concept relies on latents maintaining spatial correlations within pixels. During the diffusion process, at each step, the spatial values representing the background in the predicted latents are replaced with the corresponding spatial values from the source video latents, obtained during the inversion process. Results in Fig.~\ref{fig:personlization_results}, Fig.~\ref{fig:crossdomain_appx}, and Fig.~\ref{fig:structure_appx} utilize latent blending to preserve the background.

\begin{table*}[b]
\centering
\begin{tabular}{lcccc}
\toprule
\multirow{2}{*}{Method} & \multicolumn{2}{c}{Structure} & \multicolumn{2}{c}{Cross-Domain} \\
 & \multicolumn{1}{l}{Image-Text} & \multicolumn{1}{l}{Image-Image} & \multicolumn{1}{l}{Image-Text} & \multicolumn{1}{l}{Image-Image} \\\\[-2ex]
\hline
 \\[-2ex]
Tune-A-Video & 25.64 & \textbf{97.74} & 25.57 & 95.01 \\
FateZero & 25.55 & 97.39 & 24.25 & 94.60 \\
VideoSwap & 26.70 & 97.71 & 27.19 & 95.13 \\
VidMP3 & \textbf{26.74} & 97.58 & \textbf{30.75} & \textbf{97.94}\\\hline
\end{tabular}
\caption{\textbf{Quantitative evaluation with CLIP ViT-L/14@336px.}}
\label{tab:clip_score}
\end{table*}

\begin{figure*}[t]
    \centering
    \includegraphics[width=0.94\linewidth]{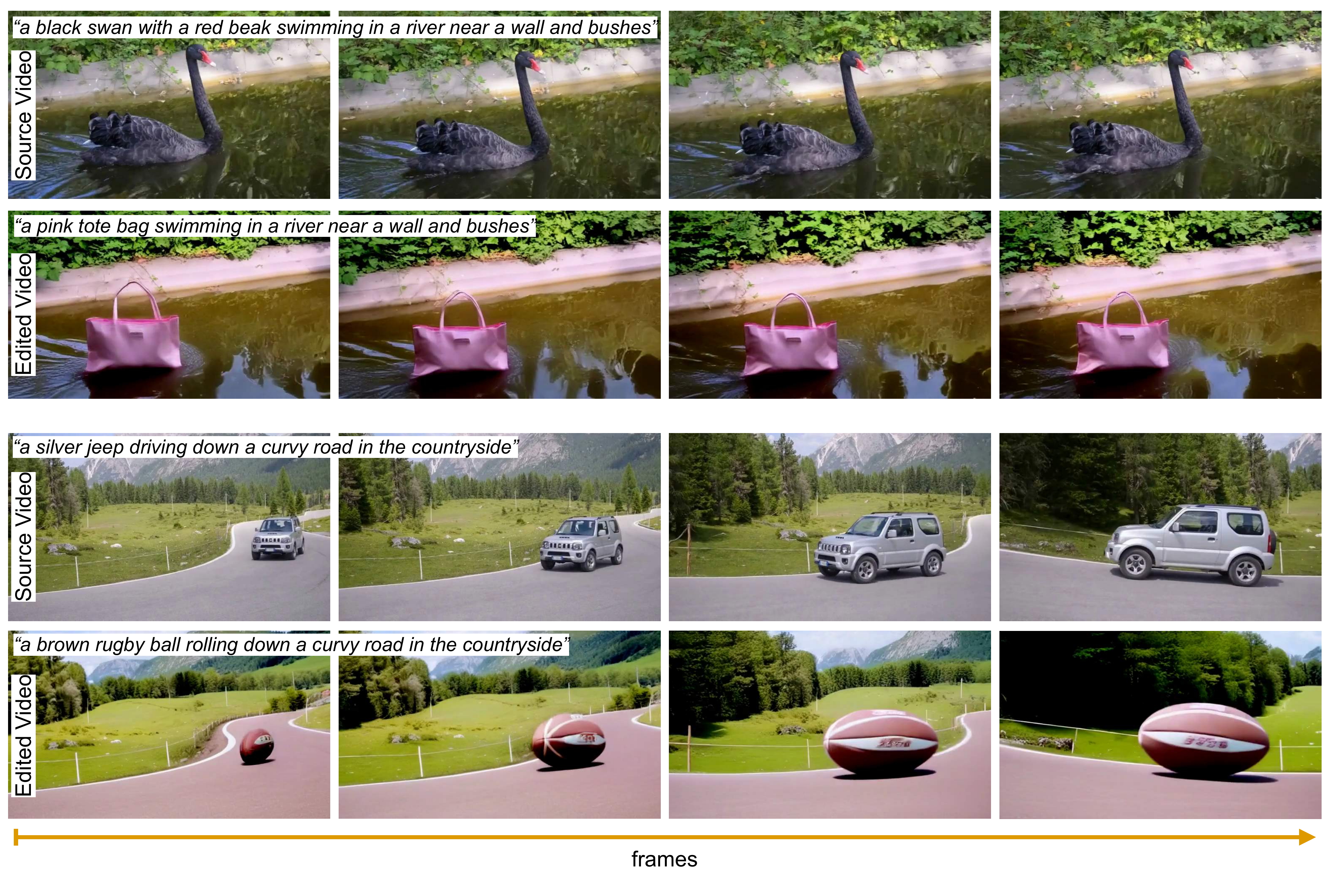}
    \caption{\textbf{More SDXL Results.} Additional examples of novel concepts generated using SDXL as the foundation model in \textsc{VidMP3}.}
    \label{fig:sdxl_appx}
\end{figure*}

 \begin{figure*}[b]
    \centering
    \includegraphics[width=1.0\linewidth]{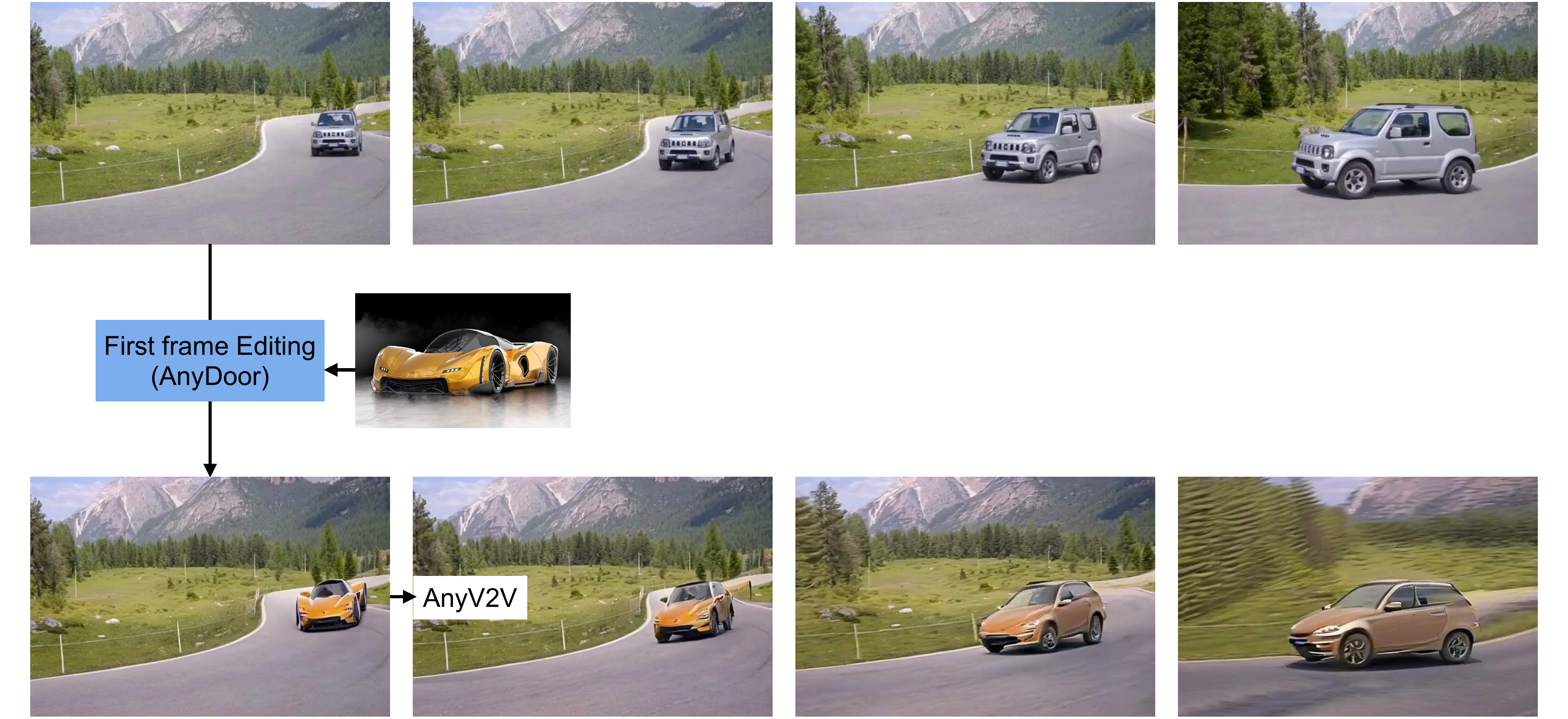}
    \caption{\textbf{Results of AnyV2V on subject swapping.} We observe that AnyV2V cannot handle pose changes in the subject. Additionally, it relies on the image editing quality of the first frame which is of poor quality and also requires human effort. Compare to Fig.8 which shows our results for the same edit.}
    \label{fig:anyv2v}
\end{figure*}

\begin{figure*}[t]
    \centering
    \includegraphics[width=1.0\linewidth]{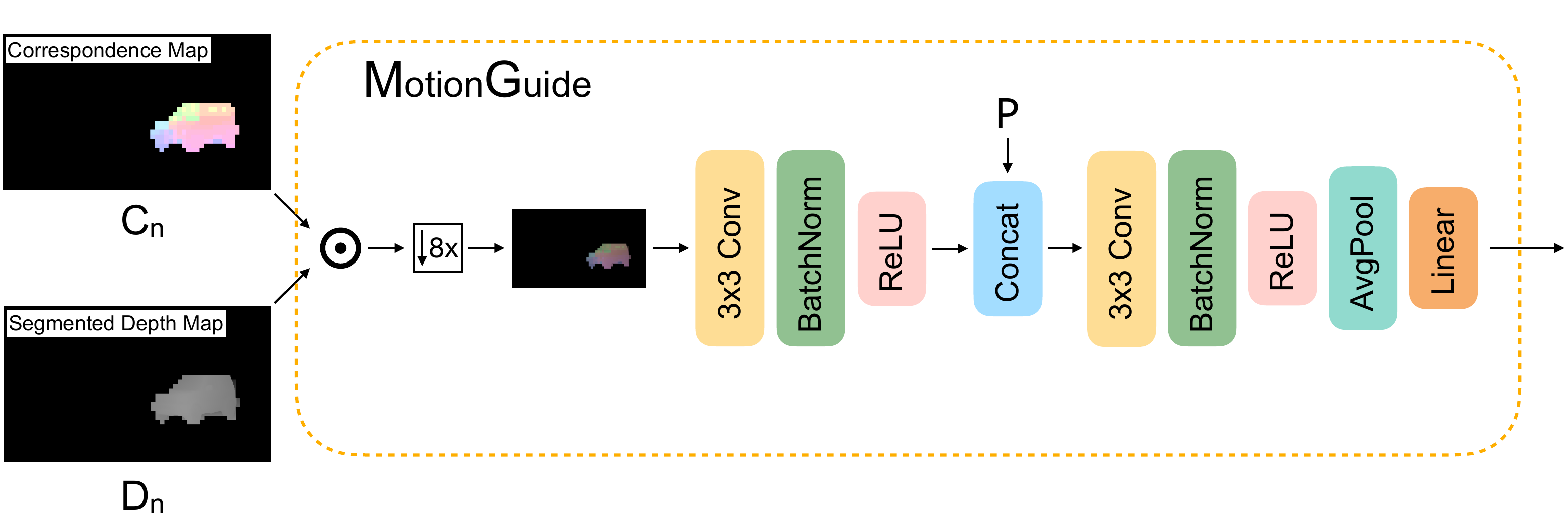}
    \caption{\textbf{\textsc{MotionGuide} Architecture.} }
    \label{fig:motionGuide_Arch}
\end{figure*}

\section{Additional results}

We provide additional results of 1) Cross-Domain Editing, 2) Structure Editing, and 3) scaling to SDXL in Figs.~\ref{fig:crossdomain_appx}, ~\ref{fig:structure_appx}, and ~\ref{fig:sdxl_appx}, respectively. We also provide some video results in the supplementary zip file.

\section{First Frame Editing method}
We show results of the first frame edit propagation method AnyV2V for the case of swapping ``silver jeep" to a novel car whose image has been provided. We use AnyDoor to edit the first frame of the video (as AnyV2V suggests), and then provide this to AnyV2V for video editing. Firstly, we note that the editing quality of AnyDoor is subpar and also requires human effort in masking regions. However, it should be noted that the quality of AnyDoor is better for editing than the prompt-based method InstructPix2Pix which is another option employed by AnyV2V for first frame editing. Secondly, we notice that the identity of the car changes drastically over frames finally becoming gray, which shows that AnyV2V cannot handle pose changes of objects. For comparison with \textsc{VidMP3} for this case, please refer to the first row of Fig.~\ref{fig:personlization_results}.

\section{\textsc{MotionGuide} Architecture}

We utilize the correspondence map $C_n$ and segmented depth map $D_n$ as inputs to the \textsc{MotionGuide} module. To reduce computational complexity, these inputs are scaled down using the same scaling factor applied by the VAE encoder in most T2I models. The first convolutional layer then expands the input from three channels to 64 channels. The second convolutional layer operates on a 128-channel input, formed by concatenating the positional encoding $P$ with the output of the previous convolutional layer, producing an output of 256 channels. Finally, a linear layer transforms this 256-channel activation into the desired number of channels, making it suitable for integration with the attention layer's values. Figure~\ref{fig:motionGuide_Arch} illustrates the architecture of the \textsc{MotionGuide} module.


\section{Evaluation}
We evaluate \textsc{VidMP3} and previous methods using the same videos as described in Sec.\ref{sec:experiments} \textbf{Datasets}. 180 edited results from each video editing method are compared in both automatic and human evaluation settings as described below.

\textbf{Automatic Evaluation.}

We utilized CLIP-Score~\cite{hessel2021clipscore} as an automatic evaluation metric to quantitatively assess all video editing methods. To compute the video-text alignment score for a test video, we averaged the image-text alignment scores across all its frames. Subsequently, the video-text alignment scores of all test videos were averaged to derive the overall video-text alignment score for each method. 

As a preliminary analysis of temporal consistency in a test video, we calculate the image-to-image alignment score for every alternate frame pair and average these scores across all frames to determine the video's temporal consistency. The temporal consistency scores of all test videos are then averaged to compute the overall temporal consistency score for each method. 

The results of the automatic evaluation, categorized into (1) Structure Editing and (2) Cross-Domain Editing, are summarized in Table~\ref{tab:clip_score}. For structure editing, \textsc{VidMP3} achieves performance comparable to previous methods. However, for cross-domain editing, \textsc{VidMP3} demonstrates significantly superior performance.

\textbf{Human Evaluation.} We conducted a controlled laboratory study to evaluate different methods based on the following criteria: (1) Subject Identity, (2) Motion Alignment, (3) Temporal Consistency, and (4) Overall Preference. Preference-based feedback was collected for all 180 edits from 10 participants, with each participant providing ratings for all edits, resulting in a total of 180 ratings per participant. While a larger sample size of feedback per edit is generally preferred, the task of identifying issues in the edited results is relatively straightforward, making 10 participants a reasonably sufficient number for this study. The human evaluation results shown in Fig.~\ref{fig:humanStudy} of the main paper clearly indicate a strong preference for our method.

For each editing concept, the rating interface displays the source video, source prompt, edit prompt, and the edited videos generated by all methods under comparison. For personalized video editing, the interface also includes the reference images utilized for ED-LORA-based personalization.

\end{document}